\documentclass[a4paper,10pt]{article}
\usepackage[utf8]{inputenc}
\usepackage[a4paper,left=2.5cm,right=2.5cm,top=2.5cm,bottom=2.5cm]{geometry}
\usepackage{natbib}
\usepackage[hidelinks]{hyperref}
\usepackage{graphicx}
\usepackage{bbm}
\usepackage{amsmath}
\usepackage{array}
\usepackage[font=small,labelfont=bf]{caption}

\renewcommand{\arraystretch}{1.2}

\usepackage{fancyhdr}

\newcommand{\NEST}{\texttt{NEST}}

\title{Robust trajectory generation for robotic control on the neuromorphic research chip Loihi}
\author{Carlo Michaelis, Andrew B. Lehr and Christian Tetzlaff}
\date{November 17, 2020}

\begin{document}

\maketitle

\begin{abstract}
\noindent Neuromorphic hardware has several promising advantages compared to von Neumann architectures and is highly interesting for robot control.
However, despite the high speed and energy efficiency of neuromorphic computing, algorithms utilizing this hardware in control scenarios are still rare. 
One problem is the transition from fast spiking activity on the hardware, which acts on a timescale of a few milliseconds, to a control-relevant timescale on the order of hundreds of milliseconds.
Another problem is the execution of complex trajectories, which requires spiking activity to contain sufficient variability, while at the same time, for reliable performance, network dynamics must be adequately robust against noise.
In this study we exploit a recently developed biologically-inspired spiking neural network model, the so-called anisotropic network. We identified and transferred the core principles of the anisotropic network to neuromorphic hardware using Intel's neuromorphic research chip Loihi and validated the system on trajectories from a motor-control task performed by a robot arm.
We developed a network architecture including the anisotropic network and a pooling layer which allows fast spike read-out from the chip and performs an inherent regularization.
With this, we show that the anisotropic network on Loihi reliably encodes sequential patterns of neural activity, each representing a robotic action, and that the patterns allow the generation of multidimensional trajectories on control-relevant timescales.
Taken together, our study presents a new algorithm that allows the generation of complex robotic movements as a building block for robotic control using state of the art neuromorphic hardware.
\end{abstract}

\section{Introduction} \label{section:introduction}

During infancy, humans acquire fine motor control, allowing flexible interaction with real world objects.
For example, most humans can effortlessly grasp a glass of water, despite variations in object shape and surroundings.
However, achieving this level of flexibility in artificial autonomous systems is a difficult problem.
To accomplish this, such a system must accurately classify inputs and take appropriate actions under noisy conditions.
Thus, increasing robustness to input noise is crucial for the development of reliable autonomous systems \citep{khalastchi2011online, naseer2018robust}.

Neuromorphic hardware is based on highly parallel bio-inspired computing, which employs decentralized neuron-like computational units.
Instead of the classical separation of processing and memory, on neuromorphic hardware information is both processed and stored in a network of these computational units.
Neuromorphic architectures offer faster and more energy-efficient computation than traditional CPUs or GPUs \citep{tang2019spiking, blouw2019benchmarking}, which is a vital feature for autonomous systems.
However, porting existing robot control algorithms (e.g. \citealp{ijspeert2002movement}) to neuromorphic hardware is per se ambitious (but see \citealp{eliasmith2004neural,dewolf2016spiking, voelker2017methods}) and difficult to optimize to the specific hardware architecture.
At the same time, the development of new algorithms is also challenging due to the decentralized design principle of neuromorphic hardware as a network of computational units \citep{lee2018training}.


The basic network type for the various neuromorphic architectures developed in recent years \citep{schemmel2010wafer, furber2014spinnaker, neckar2018braindrop, davies2018loihi} are spiking neural networks (SNNs), coined third generation neural networks \citep[for review, see][]{maass1997networks,tavanaei2019deep}. 
In particular, the reservoir computing paradigm, such as echo state networks \citep{jaeger2001echo, jaeger2007echo} or liquid state machines \citep{maass2002real}, often serves as an algorithmic basis.
In reservoir computing a randomly connected SNN provides a ``reservoir'' of diverse computations, which can be exploited by training weights from the reservoir units to additional units that constitute time-dependent outputs of the system.

The internal dynamics of the reservoir or SNN generally provide a sufficient level of variability such that arbitrary output functions on a control-relevant timescale can be read out.
However, the system fails if the input is noisy or perturbations arise while the trajectory is being performed \citep{maass2002real, sussillo2009generating, laje2013robust, hennequin2014optimal}.
That is to say, spiking dynamics in SNNs are often unstable, meaning that small changes in the initial conditions result in different spiking patterns \citep{sompolinsky1988chaos, van1996chaos, brunel2000dynamics, london2010sensitivity}.
Thus, when an output is trained using such a spiking pattern, low levels of noise lead to a deviation of the estimated output from the target output and stable trajectories can only be obtained on a timescale of milliseconds.
On the other hand, attractor dynamics provide highly stable, persistent activity \citep{amit1992modeling, tsodyks1999attractor}; however, they tend to lack the variability in the spiking dynamics required for complex output learning \citep{nachstedt2017working}.
This implies a stability-variability trade-off, also denoted as a robustness–flexibility trade-off \citep{pehlevan2018flexibility}.

\begin{figure}[!t]
    \centering
    \includegraphics[width=\textwidth]{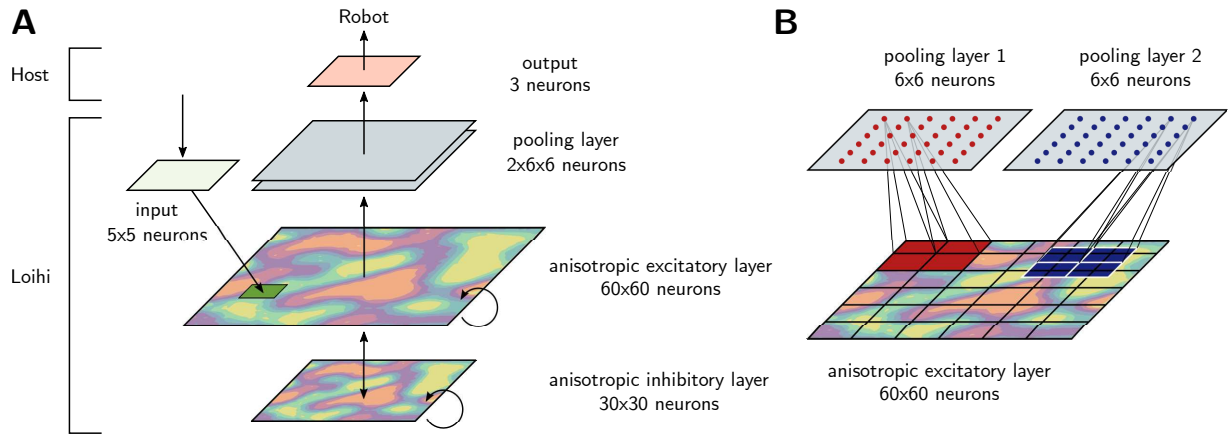}
    \caption{The architecture of the robot control algorithm. \textbf{(A)}~The anisotropic network is the basis of the architecture, indicated by the colored layers, and consists of $3600$ excitatory and $900$ inhibitory neurons. An area of $25$ excitatory neurons receive external input. The excitatory neurons of the anisotropic network project to a pooling layer. The anisotropic network, the input neurons and the $72$ pooling neurons are simulated on-chip. Finally, the spiking activity is used to train a trajectory. \textbf{(B)}~In detail, the pooling layer contains $72$ neurons organised into two $6 \times 6$ grids. Each pooling neuron receives all-to-one feed-forward connections from a $10 \times 10$ square patch of the anisotropic network's excitatory layer. Hence, the pooling layer has $10 \times 10$ window size and stride $5$.}
    \label{fig:1}
\end{figure}


A number of approaches have been developed in recent years to stabilize the spiking dynamics of SNNs while retaining sufficient variability for output learning \citep{laje2013robust, hennequin2014optimal, pehlevan2018flexibility, vincent2020learning}.
To improve stability, recent approaches used feed-forward structures \citep{pehlevan2018flexibility} or employed supervised learning rules \citep{laje2013robust}.
While feed-forward structures provide stable activity patterns, in general these play out on a very fast timescale \citep{Zheng2014Robust} or require neural/synaptic adaptation such that activity moves between neuron groups \citep{york2009recurrent, itskov2011cell, murray2017learning, maes2020learning}. 
And since for supervised learning all states in the network need to be accessible at each computing unit, these so-called global learning rules are not compatible with most neuromorphic hardware.

 
Thus, achieving stable activity patterns on a control-relevant timescale in a network architecture and learning regime capable of running on neuromorphic hardware remains an open problem.
Necessary criteria are that (1) learning or adaptation mechanisms in the SNN should be local to individual synapses, or synapses should be static, (2) sequential activity patterns should remain active for hundreds of milliseconds, (3) spike patterns should contain sufficient variability for arbitrary output learning, and (4) the network should possess noise-robust neuronal dynamics.
Meeting these criteria is especially difficult for recurrent network structures, like reservoir networks.
However, the so-called anisotropic network model appears to be a promising candidate \citep{spreizer2019space}. 
The model is based on a biologically-inspired rule for forming spatially asymmetric non-plastic connections.
Thus, synapses are static, meeting the first criterion, and the timescale of activity sequences is on the order of tens to hundreds of milliseconds, fulfilling the second criterion.
However, whether the model also fulfills the third and fourth criteria, sufficient variability and stability under input noise, has not yet been assessed.


In this paper we use the anisotropic network as a building block for a novel algorithm yielding robust robotic control.
We implement the network architecture on Kapoho Bay, a neuromorphic hardware system from Intel containing two Loihi chips \citep{davies2018loihi}, and show that this approach can be used to learn complex trajectories under noisy input conditions on a control-relevant timescale.
Furthermore, we demonstrate that this neuromorphic network architecture can not only robustly represent complex trajectories, but even generalize beyond its training experience.

\section{Methods} \label{section:methods}

We first describe the architecture of the novel algorithm implemented on the neuromorphic chip Loihi, which supports robust robotic control of movement trajectories.
The anisotropic network and its implementation is then explained in detail.
Finally analyses methods to evaluate the implementation of the anisotropic network on Loihi, the stability of its network dynamics, and the learning of complex movement trajectories are described.

\subsection{Architecture of the algorithm for robotic control}

The architecture, shown in Figure \ref{fig:1}A, was designed to support the storage and execution of stable movement trajectories in real-time.
The architecture consists of an input layer, an anisotropic network layer, and a pooling layer, all of which are fully implemented on Loihi.
Spike patterns from the anisotropic network or the pooling layer are read out and serve as the basis for training output units.

The basic computational structure for the robotic control algorithm is the anisotropic network.
Excitatory and inhibitory neurons are initialized with local, spatially inhomogeneous connections as described below.
An input is connected to a grid of $5\times 5$ excitatory neurons in order to start spiking activity with a short input pulse. 

The excitatory neurons of the anisotropic network are connected to a pooling layer with $72$ excitatory neurons.
Pooling layer neurons are organised into two grids with a size of $6 \times 6$ neurons, as shown in Figure \ref{fig:1}B.
Each neuron in the pooling layer receives input from a $10 \times 10$ group of excitatory neurons from the anisotropic network.
These projections are all-to-one and all feed-feedforward weights are equal.
In other words, the pooling layer has $10 \times 10$ window size and stride $5$.

Depending on the task, either the excitatory neurons of the anisotropic network or the $72$ neurons of the pooling layer are read out.
Since reading out data from Loihi is a bottle neck that reduces the simulation speed considerably, the pooling layer is designed to reduce read out and therefore increase simulation speed.
Finally, linear regression is applied to the spiking activity of the read-out (see section \nameref{subsection:stability}).

\subsection{The anisotropic network}

We briefly describe the main principles of the anisotropic network.
For an in depth treatment, we refer readers to \cite{spreizer2019space}.

In locally connected random networks (LCRNs), neurons are distributed in (connectivity) space (e.g. on a 2D grid or torus) and the connection probability between two neurons decreases (possibly non-monotonically) with the distance between them. 
Stable bumps of spatially localised activity can arise in LCRNs \citep{roxin2005role, hutt2008local, spreizer2017activity, spreizer2019space} and these activity bumps can move through the network in a stream-like manner if spatial asymmetries are introduced into the local connectivity \citep{spreizer2019space}.

\begin{figure}[!t]
    \centering
    \includegraphics[width=\textwidth]{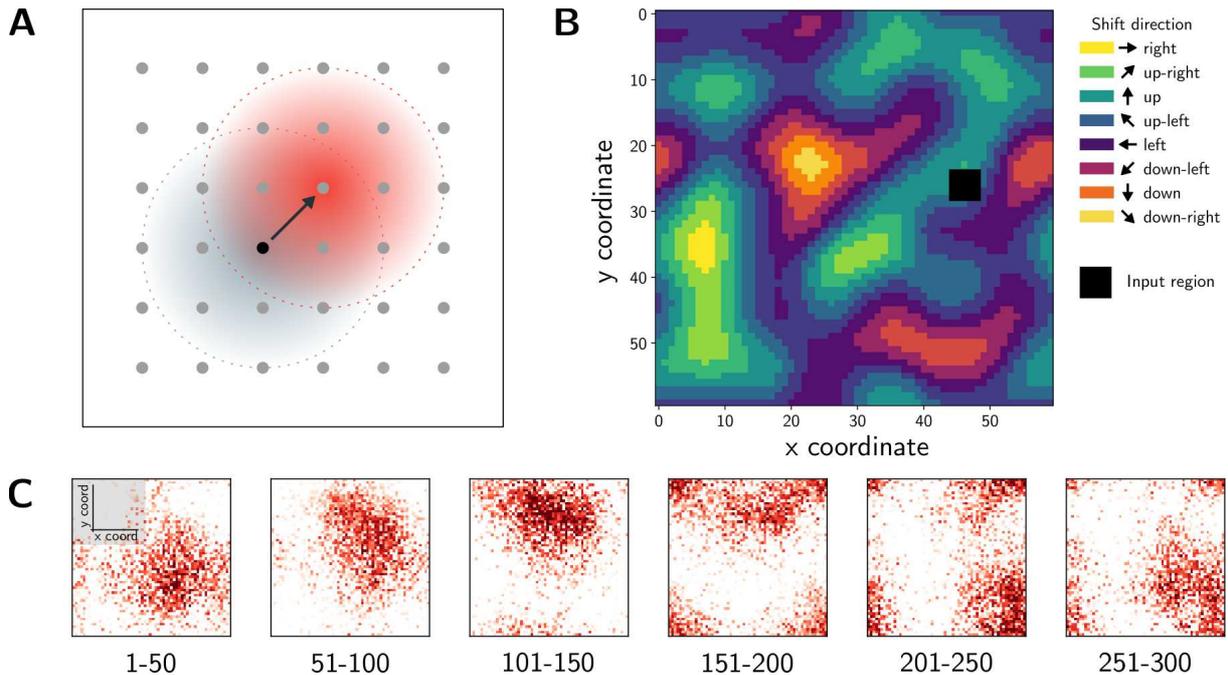}
    \caption{The connectivity underlying the anisotropic network and resulting network dynamics. \textbf{(A)}~Neurons are evenly distributed on a 2D grid, folded to form a torus. The black dot represents a reference neuron. From here outgoing connections can be drawn locally with a Gaussian distribution, either symmetrically (grey-blue) or non-symmetrically (red). In the non-symmetrical case the center of the Gaussian distribution is shifted by one neuron. This shift can be chosen in different ways. Either with the same shift direction for every neuron (homogeneously), a random shift direction for every neuron, or with a specific distribution. \textbf{(B)}~If for each neuron the shift direction is chosen based on Perlin noise, the distribution of shift directions for the whole network results in an \emph{anisotropic} connectivity structure. The black region represents spatially-localised input. \textbf{(C)}~Given local input, the anisotropic network forms a bump of activity, which moves in a stream-like pattern through the inhomogeneous structure of connections. Here, we used the Loihi implementation and binned the spikes into non-overlapping time windows. Each graph shows the average firing rate over $50$ time steps, color code depicts firing rate.
   }
    \label{fig:2}
\end{figure}

The anisotropic EI-network consists of both excitatory and inhibitory neurons arranged on a 2D torus.
Neurons project their axons in a distance dependent way with connection probability decreasing monotonically according to a Gaussian distribution.  
In a standard LCRN, axon projection profiles are centered at the neuron and axons project symmetrically in all directions.
In the anisotropic EI-network, the Gaussian distribution is shifted for excitatory neurons such that connections to other excitatory neurons are formed preferentially in a particular direction (Figure \ref{fig:2}A).

A so-called landscape is computed on the torus using Perlin noise \citep{perlin1985image}, and each point on the grid (neuron) is assigned a direction based on this. 
The Perlin landscape ensures that the preferred direction of nearby neurons are similar while preferred directions of those far apart are uncorrelated (Figure \ref{fig:2}B).
Each excitatory neuron's connectivity profile is shifted by one grid point in its preferred direction, resulting in spatially asymmetric but correlated connectivity.
When a set of neurons in close proximity are stimulated, spatio-temporal sequences of activity lasting tens to hundreds of milliseconds are elicited (Figure \ref{fig:2}C).

Taken together, a biologically plausible rule can generate spatially asymmetric connectivity structures supporting spatio-temporal sequences.
\cite{spreizer2019space} show that if (1)~individual neurons project a small fraction ($\sim$2-5\%) of their axons preferentially in a specific direction (Figure \ref{fig:2}A), and (2)~neighboring neurons prefer similar directions (Figure \ref{fig:2}B), then sequences of neural activity propagate through the network (Figure \ref{fig:2}C).
This simple generative connectivity rule results in feed-forward paths through the otherwise locally connected random network.

\subsection{Anisotropic network implementation}

We adapted the anisotropic EI-network model from \cite{spreizer2019space}.
Since the total number of connections currently supported by the Loihi \texttt{NxSDK}-API is limited (see \nameref{section:discussion}), it was necessary to reduce network size by a factor of four to $npop_E = 3600$ and $npop_I = 900$.
Each neuron projects to $p_{conn} \times npop_E = 180$ excitatory targets and $p_{conn} \times npop_I = 45$ inhibitory targets, where $p_{conn} = 0.05$ is the connection probability.
Connection probability decreases with distance according to a Gaussian distribution with space constants given in Table S1.
We first adapted the anisotropic EI-network model within \NEST~and then transferred it to Loihi, tuning the network to qualitatively match the behavior of the \NEST~simulation.

\paragraph*{\NEST~implementation.}


Neurons were modelled as leaky integrate-and-fire (LIF) neurons, with sub-threshold membrane potential~$v$ of neuron $i$ evolving according to:

\begin{equation}\label{eq:nestlif}
    C_m \frac{dv_i}{dt} = -g_L (v_i(t) - E_L) + I_i(t) + I^{input}_{i}(t),
\end{equation}

\noindent where $C_m$ is the membrane capacitance, $g_L$ the leak conductance, and $E_L$ the reversal potential.
For neuron~$i$, $I_i(t)$ is the total synaptic current from its recurrent connections and $I^{input}_i(t)$ the current induced by external input.

The total synaptic current $I_i(t)$ to neuron $i$ at its recurrent synapses is the sum of the current transients at each of its synapses, $I_i(t) = \sum_{j}I_{ij}(t)$.
When a pre-synaptic neuron spikes, a current transient is elicited with temporal profile given by an alpha function:

\begin{equation}\label{eq:alpha}
    I_{ij}(t) = J^{syn} \frac{t - t_{j,k}}{\tau_{syn}} \, \text{exp} \bigg(-\frac{t - t_{j,k}}{\tau_{syn}} \bigg).
\end{equation}

\noindent Note here that the superscript \textit{syn} can denote both excitatory (\textit{exc}) and inhibitory (\textit{inh}) synapses.
Synaptic strength is $J^{syn}$, synaptic time constant is $\tau_{syn}$, and spike time is $t_{j,k}$ for the $k^{\text{th}}$ spike from neuron $j$.

To compensate for the decreased network size and hence fewer recurrent connections (see above) we scaled up the synaptic weights.
The excitatory synaptic current was scaled up by a factor of four to $J^{exc} = 40.0 \, pA$.
To ensure persistent spiking activity in response to an input pulse, the ratio of recurrent inhibition and excitation was reduced to $g = 4$.
As a result, $J^{inh} = -g \times J^{exc} = -160.0 \, pA$.

Activity was triggered by external input to a subset of neighbouring neurons, each of which receives an input pulse of $500$ spikes with synaptic strength $J^{input} = 1.0 \, pA$ arriving according to a Gaussian distribution with standard deviation of $1 \, ms$. 

\paragraph*{Loihi implementation.}

For the implementation on neuromorphic hardware we used the research chip Loihi from Intel \citep{davies2018loihi}, which is a digital and deterministic chip that is based on an asynchronous design.
The board we used contains two chips, with each chip providing $128$ neuron cores and three embedded x86 CPUs.
Each neuron core time-multiplexes the calculation and allows the implementation of up to $1024$ neurons each.
We distributed the total of $4572$ utilized neurons (reservoir and pooling layer) with $20$ neurons per core.
Computation on the chip is performed in discrete time steps and has no relation to physical time.
Finally, the Loihi board is connected to a desktop computer, called host in the following, via a serial bus (USB).

We translated the \NEST~implementation to the \texttt{NxSDK} (version 0.9.5-daily-20191223) for Loihi, provided by Intel labs \citep{lin2018programming}.
For this, we developed a software framework \texttt{PeleNet}\footnote{\url{https://github.com/sagacitysite/pelenet/tree/neurorobotics}}, based on the \texttt{NxSDK}, especially for reservoir networks on Loihi.
This framework was used for all simulations in this study.

The Loihi chip implements a leaky integrate-and-fire (LIF) neuron with current-based synapses and the membrane potential $v$ of neuron $i$ evolves according to

\begin{equation}\label{eq:voltage}
    \frac{dv_i}{dt} = - \tau_v^{-1} v_i(t) + I_i(t) + I^{input}_i(t) - v_{th} \sigma_i(t),
\end{equation}

\noindent where $\tau_v$ describes the time constant, $v_{th}$ the firing threshold, $I_i(t)$ the total synaptic current from recurrent connections, $I^{input}_i(t)$ the current induced by the input, and $\sigma_i(t)$ denotes whether neuron $i$ spiked at time~$t$.
The first term on the right-hand side controls voltage decay, the second/third term increases the voltage according to the synaptic/input currents, and the last term resets the membrane potential after a spike occurs.

While the \NEST~implementation uses alpha-function shaped synaptic currents (see Equations \ref{eq:nestlif} and \ref{eq:alpha}), Loihi's current-based synapses implement instantaneous rise and exponential decay.
The total synaptic current from recurrent connections to neuron $i$ is given by

\begin{equation}\label{eq:current-loihi}
    I_i(t) = \sum_{i \neq j} J^{syn}_{ij} (\alpha_I * \sigma_j)(t) + I_i^{bias},
\end{equation}

\noindent where $J^{syn}_{ij}$ is the synaptic strength from neuron $j$ to neuron $i$ which can be excitatory ($J^{exc}$) or inhibitory ($J^{inh}$) and $I_i^{bias}$ is a bias term. The $\sigma_j(t)$ represents the incoming spike train from neuron $j$ and $\alpha_I(t)$ a synaptic filter.
The spike train for a neuron $j$ is given by a sum of Dirac delta functions with

\begin{equation}\label{eq:sigma}
    \sigma_j(t) = \sum_k \delta(t - t_{j,k}),
\end{equation}

\noindent where $t_{j,k}$ is the time of spike $k$ for neuron $j$. 
The function simply indicates whether neuron $j$ spiked in time step $t$.
The spike train is convolved with a synaptic filter given by

\begin{equation}\label{eq:alphau}
    \alpha_I(t) = \tau_I^{-1} \exp\left(-\frac{t}{\tau_I}\right)\,H(t),
\end{equation}

\noindent where $\tau_I$ is a time constant and $H(t)$ the unit step function.

With Equations~\ref{eq:sigma} and~\ref{eq:alphau} we can bring Equation~\ref{eq:current-loihi} into a form which is comparable to Equation~\ref{eq:alpha}. 
Setting $I_{bias} = 0$, we get

\begin{alignat}{2}\label{eq:current}
    I_i(t) &=& &\sum_{i \neq j} J^{syn}_{ij} (\alpha_I * \sigma_j)(t) \nonumber\\
    &\overset{\text{Eq.}(\ref{eq:sigma})}{=}& &\sum_{i \neq j} J^{syn}_{ij} \sum_x \alpha_I(x) \sum_k \delta((t - x) - t_{j,k})) \nonumber\\
    &=& &\sum_{i \neq j} J^{syn}_{ij} \sum_k \alpha_I(t - t_{j,k})  \nonumber\\
    &\overset{\text{Eq.}(\ref{eq:alphau})}{=}& &\sum_{i \neq j} J^{syn}_{ij} \sum_k \tau_I^{-1} \exp\left(\frac{t_{j,k} - t}{\tau_I}\right)H(t-t_{j,k}).
\end{alignat}

\noindent Due to the filter, the input current induced by a pre-synaptic spike decays exponentially for each following time step.
And instead of rising slowly, at the time of a spike, $t = t_{j,k}$, synaptic current increases by $\tau_I^{-1} J^{syn}_{ij}$.
Thus, compared to the neuron model from the anisotropic network implementation in \NEST~\citep{spreizer2019space}, the hardware-implemented neuron model on Loihi differs since it lacks a current rise time.

\subsection{Comparing the implementations}

Network activity was started with the input mentioned above and $500$ discrete time steps in Loihi and $500\,ms$ in \NEST~were recorded.
In \NEST~the resolution was set to $dt = 0.1\,ms$ (see also Table S1) per simulation step, while in Loihi a physical time is not defined.
After the simulation, the \NEST~spiking data was binned to $1\,ms$ to match the Loihi data.
Note that, given the refractory period of $2\,ms$, the binned spike trains still contain binary values, but with a less precise information about the sub-millisecond spike times.
In the end, both data sets, the spike trains for \NEST~and the spike trains for Loihi contained $500$ discrete steps.

To compare the spiking patterns between \NEST~and Loihi quantitatively, we calculated the mean firing rate of groups of excitatory neurons in both networks, which is shown in Figure \ref{fig:3}B.
For this, we split the two dimensional network topology into a $6\times6$ grid, analogous to the grid used for the pooling layer (see Figure \ref{fig:1}B) such that each grid position represents a group of $100$ neurons.
The indices of the groups are chosen from top left to bottom right.
For each group, we averaged the firing rate over the $500$ time steps resulting in $36$ values. 

\subsection{Stability and output learning} \label{subsection:stability}

To analyse the stability (Figure \ref{fig:5}) of the network and for the output learning (Figure \ref{fig:6}), we applied another protocol.
Note that from this point on, the \NEST~implementation was not used.
Out of the $25$ excitatory neurons connected to the input, we stimulated only $24$ neurons such that $1$ neuron stays silent (Figure \ref{fig:4}A).
This input grid then allows $25$ different input configurations and therefore $25$ different trials with a noise level of $4\%$.
Every trial was recorded for $215$ time steps and then the activity was stopped by resetting the membrane voltages.
We did this by applying a \texttt{C} code that runs on one x86 core on each chip (a so called SNIP).
After waiting $30$ time steps, the next input was applied to the network.
The applied protocol is indicated by arrows in Figure \ref{fig:5}A on top of the spike train plots.

To learn trajectories from the spike patterns we used multiple linear regression, which was applied to two different tasks.
In the \textit{representation task}, we estimated model parameters based on all $25$ trials and tested on one of them.
In the \textit{generalisation task}, training was performed on only $24$ trials and testing was done using the remaining trial.
Both tasks are sketched in Figure \ref{fig:4}B.
To compare the anisotropic network with a classical reservoir computing approach, we also implemented a randomly connected network on Loihi and exchanged the anisotropic network in our network architecture with a randomly connected network of equal size.
We set the parameters of the random network such that the main statistics of both networks match.
The firing rate in both networks is in a range of $0.1-0.2$ spikes per number of neurons (Figure \ref{fig:5}A bottom) and the mean Fano factor over all trials is relatively similar with $\overline{FF}_{\text{rand}} = 0.80 \pm 0.01$ for the randomly connected network and $\overline{FF}_{\text{aniso}} = 0.84 \pm 0.002$ for the anisotropic network.

After all data were recorded, in a first step, we prepared the data for the estimation of the regression model.
Due to the slow rise in the firing rate of the network (Figure \ref{fig:3}C), the very first time steps contain little information.
Therefore, we omitted the first $5$ time steps which reduces the length of the data set to $210$ per trial.
Before the linear regression was applied to the spike data, the spike trains were binned in order to smooth our spiking data.
We used a sliding window with a width of $10$ time steps, which reduced the length of the data set again from $210$ to $200$.

Next we used the binned data of the $200$ time steps to estimate the regression parameters.
In addition to the spiking data from the neurons, an intercept was added such that the number of parameters equal the number of neurons plus one.
The linear regression model was performed on the CPU of the host computer, using the spiking data from the readout provided by Loihi.
The two different tasks, the representation task and the generalisation task, were performed using the spiking data from the anisotropic network as well as those from the randomly connected network.
Furthermore, we estimated output weights based on either the pooling layer neurons ($72$ neurons) or the excitatory neurons of the reservoir ($3600$ neurons), see also Figure \ref{fig:1}.
The excitatory neuron readout serves as a control and compares the pooling layer approach to a traditional readout.

For the estimation based on all excitatory reservoir neurons, we applied an elastic net regularization \citep{zou2005regularization} to avoid overfitting, due to the numerous parameters.
This regularization approach for regression models simply combines LASSO and ridge regression.
We used the \texttt{fit\_regularized} function from the \texttt{statsmodels} package in \texttt{Python}, which applies elastic net as
\begin{equation}
    \hat{\beta} = \underset{\beta}{\operatorname{argmin}} \left[\| y-X \beta \|_2^2 + \alpha\left((1-\lambda) \|\beta\|_2^2 + \lambda \|\beta\|_1\right) \right].
\end{equation}
In this variant the parameter $\alpha$ determines the degree of regularization and $\lambda$ balances between LASSO ($L1$ regularization) and ridge regression ($L2$ regularization).

To better compare the predicted function with the target function, we applied a Savitzky-Golay filter \citep{savitzky1964smoothing} to smooth the predicted function.
For this we used the \texttt{savgol\_filter} function of the \texttt{Python} package \texttt{scipy}.
We chose a window length of $21$ and an order of the polynomial of $1$ as parameters for smoothing.

We trained our algorithm on $7$ different trajectories performing ordinary robotic tasks.
The tasks are \emph{hide}, \emph{unhide}, \emph{move down}, \emph{move up}, \emph{pick and place}, \emph{put on top} and \emph{take down} (see e.g. \citealp{worgotter2020humans}).
Movement data is given in $3$ dimensional Cartesian coordinates, resulting in three outputs or -- biologically speaking -- in three rate coded output neurons.

\section{Results}\label{section:results}

We start by demonstrating that the main principles of the anisotropic network are preserved by the Loihi implementation and then confirm that the Loihi-based anisotropic network admits noise-robust spiking dynamics.
Based on these findings, we demonstrate that our architecture can learn complex trajectories under noisy input conditions.

\subsection{Implementing the computer-based anisotropic network on Loihi}

\begin{figure}[!t]
    \centering
    \includegraphics[width=\textwidth]{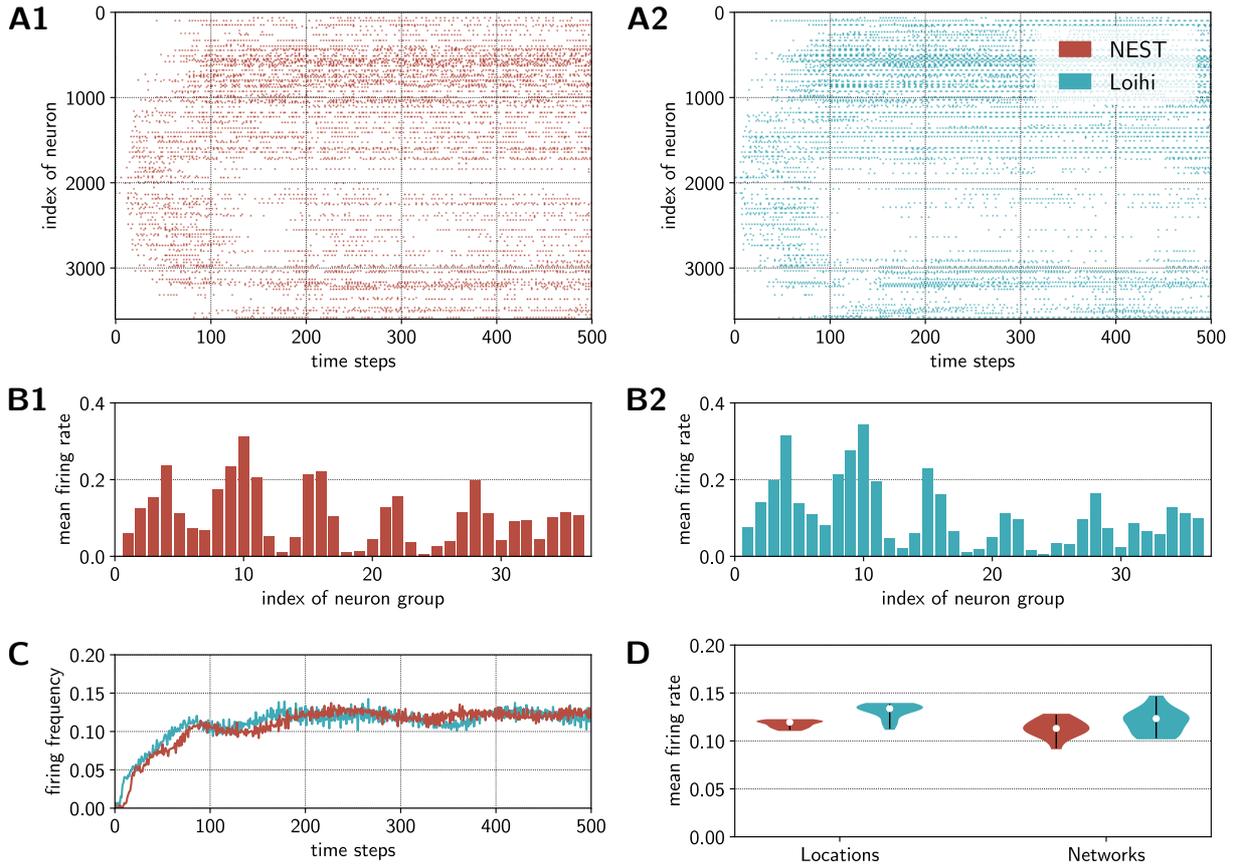}
    \caption{One example of the anisotropic network implementation on Loihi compared to activity statistics of the \NEST~simulations. \textbf{(A)}~The spiking activity is shown as spike raster plots. Red: \NEST~simulation; blue: Loihi implementation. Both networks are initialized with the same weight matrix and triggered with the same input. \textbf{(B)}~By pooling neurons into groups, each consisting of $100$ neurons, and binning across the whole simulation time, we obtain the distribution of mean firing rates (\#~spikes~/~\#~neurons~per~group)  for the \NEST~and Loihi implementation. \textbf{(C)}~The time course of the network firing rate (\#~spikes~/~\#~neurons) from the \NEST~and Loihi simulations is depicted. \textbf{(D)}~Testing one network initialization with $15$ different input locations (Locations) and one location but $15$ different network initialization (Networks) for the \NEST~as well as the Loihi implementation. White dots indicate the median of the mean firing rate (\#~spikes~/~\#~neurons) of the $15$ simulations for each case.}
    \label{fig:3}
\end{figure}

Due to the different hardware architectures, we first assess the extent to which the Loihi-based implementation of the anisotropic network agrees with the computer-based \NEST~simulation. Please note that it is not our goal to compare two neural network simulators, but to ensure that the anisotropic network implementation on Loihi preserves the main features.
For the sake of comparison, we used the same connectivity structure and input positions for both implementations.
The networks were initialised at rest and spike patterns were evoked via a spatially-localised input.
Raster plots of evoked spike trains indicate that, although the detailed spiking activity is not identical, the overall spiking pattern is mainly preserved (Figure \ref{fig:3}A). 
Accordingly, the mean firing rate of the network for each implementation evolves similarly over time (Figure \ref{fig:3}C). 

We confirmed the similarity between both implementations quantitatively, comparing the mean firing rate and firing rate variability over several input and network initializations.
Figure \ref{fig:3}D shows the distribution of mean firing rates over (1) $15$ different input positions for the the same network connectivity and over (2) $15$ different initializations of the network connectivity with a fixed input position.
Across input positions in the same network, firing rates for the Loihi implementation were $\bar f_{\text{inp}}^{\text{L}} = 0.131\pm 0.008$ and for the \NEST~implementation $\bar f_{\text{inp}}^{\text{N}} = 0.118\pm 0.004$.
Across network initializations, firing rates were $\bar f_{\text{init}}^{\text{L}} = 0.120\pm 0.013$ for Loihi and $\bar f_{\text{init}}^{\text{L}} = 0.113\pm 0.009$ for \NEST.
In addition, the ranges (minimum to maximum mean firing rate) of the obtained mean firing rates are very tight and overlap largely between both implementations.
For the locations, the values for Loihi are in a range of $0.11 \le f_{\text{inp}}^{\text{L}} \le 0.14$ and for \NEST~in an interval of $0.11 \le f_{\text{inp}}^{\text{N}} \le 0.12$.
In case of the different initialisations, we obtained mean firing rates between $0.10 \le f_{\text{init}}^{\text{L}} \le 0.15$ for Loihi and $0.09 \le f_{\text{init}}^{\text{N}} \le 0.13$ for the \NEST~implementation.
To compare the variability of the firing rate in both implementations, we evaluated the Fano factor (FF):
For different input positions, we obtained a mean of $\overline{FF}_{\text{inp}}^\text{L} = 0.83\pm 0.03$ for Loihi and $\overline{FF}_{\text{inp}}^{\text{N}} = 0.86\pm 0.01$ for \NEST.
In the case of the $15$ network initializations, the mean FF for Loihi is $\overline{FF}_{\text{init}}^\text{L} = 0.84\pm 0.02$ and $\overline{FF}_{\text{init}}^\text{N} = 0.86\pm 0.01$ for \NEST.
All FF values between Loihi and \NEST~are very close to each other and indicate that spiking is less variable than a Poisson process.

Given that the neural activity in the anisotropic network forms spatially-localised bumps moving through the network, we next measured its average spatial distribution.
For the spike rasters shown in Figure \ref{fig:3}A, we pooled the neurons into groups of $100$, taking into account the topology of the network (see Figure \ref{fig:1}B), and calculated the mean firing frequencies averaged across the whole simulation time.
This procedure provides a distribution of the mean activity across the network for both implementations (Figure \ref{fig:3}B).
Normalizing these distributions and comparing them with a Kolmogorov–Smirnov test reveals that the activity distributions from the \NEST- and Loihi-based implementations do not differ significantly ($D = 0.11$, $p = 0.97 > 0.05$).
Hence the spatial structure of activity patterns is similar in both implementations.

Taken together, we conclude that the Loihi implementation matches the \NEST-based anisotropic network implementation according to diverse statistics of the network activity.
This indicates a successful transfer of the core principles of the anisotropic network to neuromorphic hardware despite the differences in architecture.

\begin{figure}[!t]
    \centering
    \includegraphics[width=0.4\textwidth]{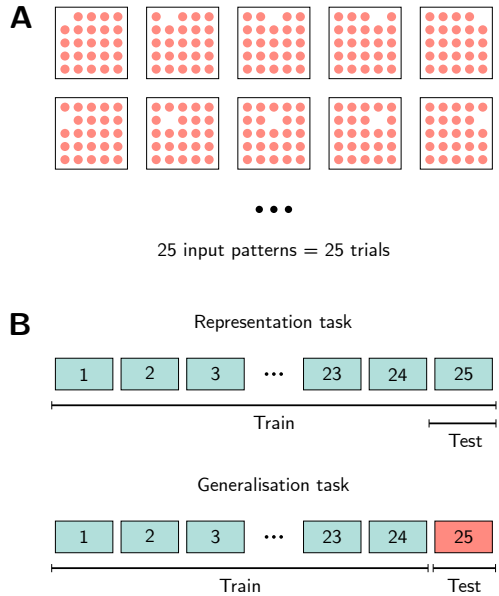}
    \caption{The input protocols for the representation and generalization task. \textbf{(A)}~A square represents an area of $25$ possible input neurons. During a trial, we triggered the network activity by activating $24$ out of the $25$. This leads to $25$ different input patterns and therefore to $25$ different trials. \textbf{(B)} We used two different training tasks. In the representation task, for the trajectory training, we used the spiking activity of all 25 trials and tested on one of them. In the generalisation task, we trained only on the spiking activity of 24 trials and tested on the spikes of a different trial, which was not used for training.}
    \label{fig:4}
\end{figure}

\subsection{The Loihi implementation of the anisotropic network is robust to input noise}

Next, to assess the robustness of the Loihi-based anisotropic network to input noise, we evaluate the stability of spiking dynamics.
An input pulse is administered to an area of the anisotropic excitatory layer consisting of $25$ neurons (Figure~\ref{fig:1}A).
In each trial, $24$ of these neurons were activated and a different neuron was systematically excluded from the input, leading to $25$ different possible input configurations and, thus, to $25$ unique trials (Figure \ref{fig:4}).

\begin{figure}[!t]
    \centering
    \includegraphics[width=\textwidth]{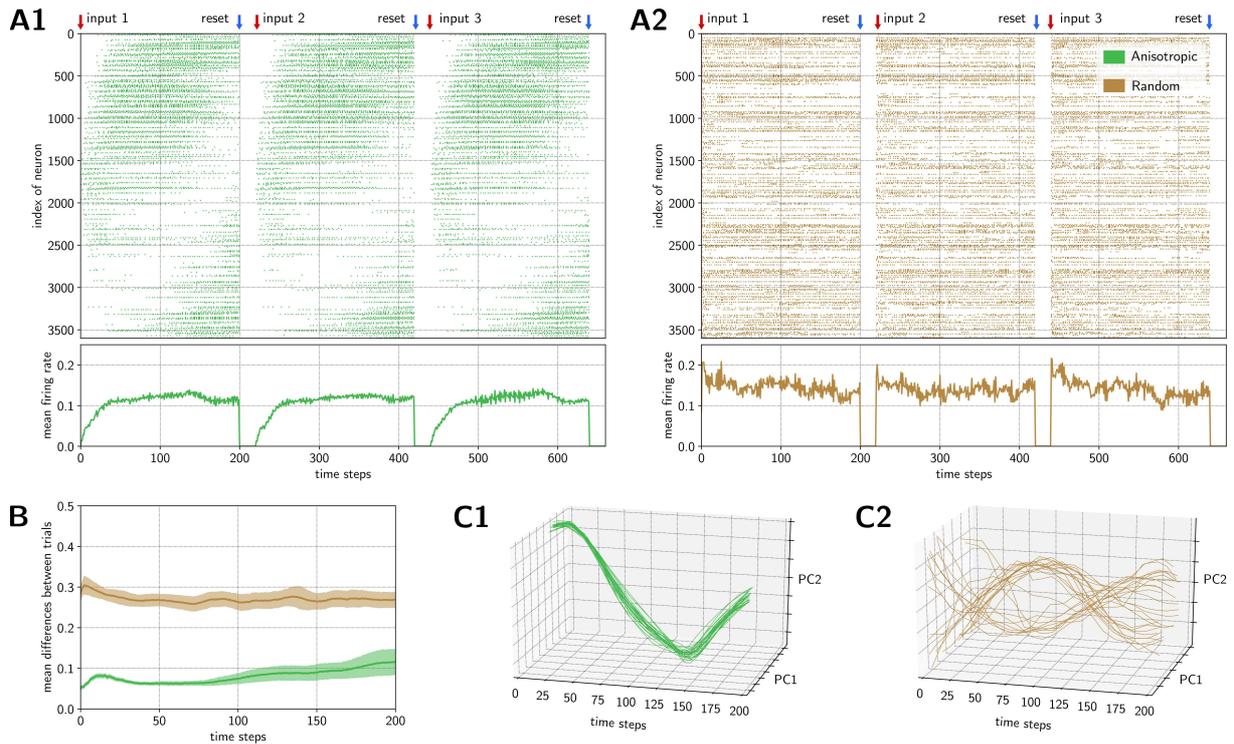}
    \caption{The Loihi implementation of the anisotropic network is robust to varying input conditions, while a randomly connected network is not. \textbf{(A)} Three examplary trials out of the 25 trials are shown for a simulation based on an anisotropic connectivity structure (green) and with randomly initialized weights (brown). The anisotropic structure clearly has a stream-like spiking pattern, where the firing rate starts slowly until it reaches a constant rate. The randomly connected network shows a Poisson-like spiking pattern, where the firing rate starts directly at a high level. \textbf{(B)}~The solid lines show the mean difference between all trial combinations and the shaded area indicates the standard deviation of the trial-to-trial differences for the anisotropic (green) and random (brown) network. \textbf{(C)} In the reduced space of the first two principal components of the activity of all 25 trials over time for both networks, we can clearly see that the spiking pattern of the anisotropic network are very similar between trials, while the activity in the randomly connected network differ much more between trials.}
    \label{fig:5}
\end{figure}

For each trial the network activity was started with a short input pulse of one time step.
We then recorded $200$ time steps of activity, stopped the activity manually and activated it again by the next input.
The protocol is also indicated on top of Figure \ref{fig:5}A.

As a control, we applied the same protocol to a randomly connected network implemented on Loihi and compared it with the anisotropic network implementation.
For this, we implemented the same algorithmic architecture, but exchanged the anisotropic network with a randomly connected network of equal size.
The spiking activity of the first three trials is shown in Figure \ref{fig:5}A1 for the anisotropic network (green) and Figure \ref{fig:5}A2 for the randomly connected network (brown).
Due to the inhomogeneous connectivity structure, the activity of the anisotropic network spreads out like a stream in the network.
Note that, given the torus network topology (see \nameref{section:methods}), the activity stream wraps around from neurons with low indices to neurons with high indices (Figure \ref{fig:5}A).
As expected, in the randomly connected network such a stream-like spread of activity does not form.

The population firing rates progress differently in the anisotropic and randomly connected networks.
The mean firing rate of the anisotropic network increases slowly until it reaches a relatively constant rate slightly above $0.1$.
The randomly connected network was tuned such that it generates a similar mean population firing rate (see \nameref{section:methods}).
However, unlike in the anisotropic network, the firing rate does not rise gradually, but instead starts at about $0.1-0.2$ straight away.
The slow start in the anisotropic network is due to the relatively small input area and the local connectivity of the network.
While moving forward in the 2D-topology, the area of activity grows step by step, which can intuitively be understood as a snowball effect.

In order to measure the stability of the spiking dynamics between different input trials, we calculated the pairwise differences between the spike patterns of all combinations of the $25$ trials.
The mean and standard deviations of these differences are shown in Figure \ref{fig:5}B for both the anisotropic network (green) and the randomly connected network (brown).
The differences between trials are much higher over the whole time course for the randomly connected network than for the anisotropic network.
For the anisotropic network, the deviations of the trial-to-trial differences are very small in the beginning and drift apart over time.
To quantify this, we performed a Levene test with three samples at time steps $10$, $100$ and $190$ which revealed that the variance stays constant between the differences of the randomly connected network trials ($W = 0.60$, $p = 0.54 > 0.05$) but increases for the anisotropic network trials over time ($W = 208.87$, $p = 3.36\cdot 10^{-75} < 0.05$).
This means that, over time, spiking patterns between some trials stay very similar whereas some trial comparisons tend to differ more.
Therefore, the anisotropic network tends to slowly diverge with time, which can also be seen by the increasing mean differences.
Importantly the mean differences in the anisotropic network remain much lower than the spiking differences between the trials in the randomly connected network, even at the end of the $200$ time steps. 
This clearly demonstrates the stabilizing feature of the anisotropic network.

To visualize differences between the single trials, we reduced the dimensionality of the spiking data by applying principal component analysis (PCA) to all trials (see \nameref{section:methods}).
The results are shown in Figure \ref{fig:5}C.
For the anisotropic network (Figure \ref{fig:5}C1), all trajectories are very similar whereas for the randomly connected network (Figure \ref{fig:5}C2) the trajectories differ considerably.
We quantified this by calculating statistics in the first dimension of the PCA space.
First, we obtained the pairwise normalized mean squared error between all trials for each network type.
The normalized mean error between the trials of the randomly connected network is $\text{MSE}_\text{rand} = 1.66$, while the anisotropic network has a mean error of only $\text{MSE}_\text{aniso} = 0.03$, which is significantly lower (Mann–Whitney U test: $U = 3572.0$, $p = 4.27\cdot 10^{-85} < 0.05$).
Even though some trajectories seem to follow a common path, in the random network, the mean standard deviation for the first principle component $\bar\sigma_{\text{rand}} = 2.50$ is significantly higher than in the anisotropic network with $\bar\sigma_{\text{aniso}} = 0.41$ (Mann–Whitney U test: $U = 0.0$, $p = 1.56\cdot 10^{-8} < 0.05$).
This indicates sufficient stability over $200$ time steps for the anisotropic network.

Taken together, this shows the ability of the anisotropic network to produce stable spiking dynamics under noisy input conditions.
In addition it confirms the successful implementation of the network on Loihi.
In the next step we will use this intrinsic stability feature of the anisotropic network to learn robust trajectories and examine if our network produces sufficient variability to learn arbitrary functions.

\subsection{Learning robust trajectories}

After having tested and demonstrated the stabilizing feature of the anisotropic network, we aimed to use its robustness to train arbitrary output trajectories.
This step makes use of the underlying network architecture shown in Figure \ref{fig:1} and adds a linear regression model on top of this architecture for a robot control task.
The overall algorithm contains the initialization, creation and simulation of the anisotropic network, which is running on the neuromorphic hardware Loihi, and the output learning of the trajectories, which is calculated on the host CPU.

To show the robustness of this algorithm, we learned $7$ different 3D-trajectories commonly used in robotic research, like pick-and-place or put-on-top (see \nameref{section:methods}).
Using these target functions, we applied two different tasks, a representation and a generalisation task, as shown in Figure \ref{fig:4}B.
In the representation task we estimated the linear regression model based on all $25$ trials and predicted one of them, showing that the variability in the anisotropic network is sufficient to learn an arbitrary function.
To show the ability of our algorithm to robustly generalise for variations in the input, we also apply a generalisation task, where we estimate the regression model on $24$ trials and predicted the trajectory for an unseen trial.

As before, here we also compare the performance of the anisotropic network with the randomly connected network as a control.
For our algorithm we estimate our model based on the $72$ pooling layer neurons, which we can read out efficiently from the chip.
Since we reduce the parameter space of the linear regression model by using only the spiking activity of the pooling layer neurons as data, we also estimated all models based on the $3600$ excitatory reservoir neurons for comparison.

\begin{figure}[!t]
    \centering
    \includegraphics[width=0.9\textwidth]{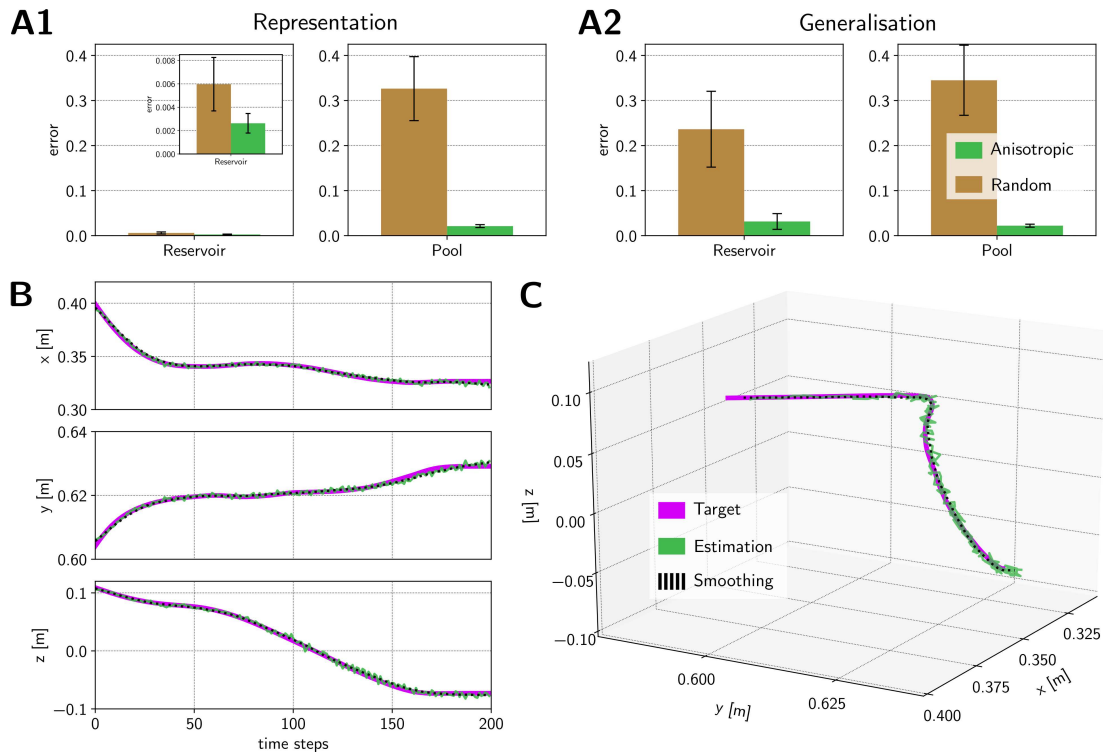}
    \caption{Predicting a trajectory on the spiking activity. \textbf{(A)}~We predicted $7$ different trajectories with $3$ dimensions each and calculated the error regarding the target trajectory using a normalized root-mean-square error. In all cases, the anisotropic network has a lower error than the randomly connected network. \textbf{(B)}~For the \textit{hide} movement, we show one example of the predicted trajectory of the anisotropic network for each of the $3$ dimensions. Here, we used the pooling neurons (green) in comparison with the target (pink) and a smoothed variant of the prediction (black dotted). The whole trajectory is shown in \textbf{(C)}.}
    \label{fig:6}
\end{figure}

Results for the representation task, based on all excitatory neurons, revealed that the excitatory reservoir neurons contain enough variability to represent an arbitrary output function with high accuracy.
An example is shown in Figure S1A1 for the randomly connected network and in Figure S1A2 for the anisotropic network.
If the model was estimated on the $72$ pooling layer neurons the number of available parameters is heavily decreased by a factor of $50$.
But still the amount of information seems to be satisfactory for the anisotropic network (Figure S1A4), but not for the randomly connected case (supplementary Figure S1A3).
Normalized root mean squared error between the predicted trajectory and the target trajectory, averaged over $7$ 3D-trajectories, are shown in Figure \ref{fig:6}A1.
The errors for all trajectories using the spiking activity of the $3600$ excitatory reservoir neurons are very low (left plot) for both networks, but interestingly even lower for the anisotropic network.
For the errors of the estimation based on the pooling layer neurons (right plot in Figure \ref{fig:6}A1), the mean error over all trajectories is still low for the anisotropic network $e_\text{aniso} = 0.02\pm 0.003$, compared to the randomly connected network $e_\text{rand} = 0.33\pm 0.07$.
Due to the inhomogeneous weight structure and the stream-like spread of spiking activity in the anisotropic network, the neurons in the pooling layer can maintain variability, as can be seen in Figure S1B1.
Intuitively, since nearby neurons have correlated activity patterns, pooling over them preserves information.
In contrast, as shown in Figure S2B2, the spiking activity in the pooling layer of the randomly connected network simply produces downsampled random spiking activity and therefore reduced variability.

In the generalisation task, the parameters for the movement trajectory were estimated based on the spiking activity of $24$ trials.
We then predicted the same trajectory based on the spiking activity elicited by a $25$th trial, not seen during training. 
This task was designed to test the robustness of the system to a variation in initial conditions.
To compare the classical reservoir computing approach with our network architecture, we trained the network based on all $3600$ neurons and on the $72$ output neurons.
In addition, this tested the ability of the pooling neurons to preserve sufficient variability while reducing the number of parameters.

For the full network read-out, we applied a linear regression model based on all excitatory neurons of the anisotropic network.
Since fitting a model based on all $3600$ neurons requires many parameters, here we used an elastic net regularization estimation method (see \nameref{section:methods} section) to reduce the number of parameters and to avoid overfitting.
Optimizing the regularization parameters resulted in $\alpha = 0.001$ and $\lambda = 0.05$.
For the pooling layer read-out, we estimated a linear regression model based on the pooling layer neurons without regularization.

In Figure \ref{fig:6}A2 we show the average normalized root-mean-squared deviation over $7$ trajectories.
In both cases (using excitatory neurons or pooling layer neurons) the error of the anisotropic network is much lower, showing that the anisotropic network has a better performance compared to a classical randomly connected network.

For the network architecture with the randomly connected network, the elastic net approach, based on all excitatory neurons, has a better performance than the linear regression approach, based on the pool neurons (t-test: $t = -4.24$, $p = 0.0001 < 0.05$).
Interestingly, the error for the anisotropic network is lower when the trajectories are estimated based on the pool neurons compared to the excitatory neurons (Mann–Whitney U test: $U = 47.0$, $p = 6.75\cdot 10^{-6} < 0.05$).
This indicates that, for the anisotropic network, the pooling layer is an equivalent, or even better regularization method compared to the elastic net approach with all excitatory neurons.

Figure \ref{fig:6}B shows all three dimensions of the predicted trajectory over time for a \emph{hide} movement.
The overall trajectory is shown in Figure \ref{fig:6}C.
We also calculated a smoothed version, using a Savitzky-Golay filter \citep{savitzky1964smoothing} (see \nameref{section:methods}), to better compare the prediction with the target.
This shows that the anisotropic network implemented on Loihi, combined with the pooling layer, contains sufficient variability to represent complex 3D trajectories while at the same time remaining stable for at least $200$ time steps.

\subsection{Simulation on Loihi in real-time}

In addition to evaluating the stability of the system, we also looked at the speed of the network simulation.
The data we used came from a Kuka robot arm, which can run fluently with an output frequency of $100\, Hz$, therefore the $200$ time steps equal $2$ seconds of movement.
In the following we denote this reference as ``real-time''.
To achieve a real-time output of spiking data from the Loihi chip, the speed of the simulation of the neurons and the data transfer from the chip to the host must be higher than the necessary data frequency of the robot for a smooth movement.

The simulation of these $200$ time steps requires $t_{3600}^\text{aniso} = 15.73\,s$ for one trial on Loihi, when all $3600$ excitatory reservoir neurons were read out from the system.
This speed is about $8$ times slower than real-time.
When reading out only from the $72$ pooling neurons, the speed increases to $t_{72}^\text{aniso} = 1.49\,s$ per trial, which is $25\%$ faster than real-time and therefore well suited for robot control.

The simulation speed of the anisotropic network ($t_{3600}^\text{aniso} = 15.73\,s$ \& $t_{72}^\text{aniso} = 1.49\,s$ per trial) and the randomly connected network ($t_{3600}^\text{rand} = 16.11\,s$ \& $t_{72}^\text{rand} = 1.73$ per trial) were nearly the same, which is expected since the number of neurons is the same and the number of synapses is similar.
Thus, the anisotropic network has, in terms of speed, no disadvantage compared to the randomly connected network.

Therefore, the pooling layer does not only reduce the sensitivity of the system but also helps to speed up the system considerably. Together, this supports robotic applications where trajectories can be stored and replayed robustly in real-time.

\section{Discussion} \label{section:discussion}


We aimed to develop an algorithm for neuromorphic hardware, which provides stable spiking dynamics under noisy input conditions, in order to make use of the low power neuromorphic chips for future autonomous systems.
For this, we derived an algorithm to store and control robotic movement sequences that unfold on a control-relevant timescale of seconds.
To validate our approach, we chose a set of $2$-second-long robot arm movements that were triggered by noisy inputs.



For our approach we chose a recently developed spiking neural network \citep{spreizer2019space} with an inhomogeneous weight structure.
In a first step we successfully transferred the main principles of this network to the Loihi research chip from Intel \citep{davies2018loihi}, a neuromorphic hardware architecture implementing spiking neurons.
In a second step we tested the stability of the anisotropic network implementation and compared its stability to a classical randomly connected network, similar to echo state networks \citep{jaeger2001echo, jaeger2007echo} or liquid state machines \citep{maass2002real}.
We finally used a pooling layer (Figure \ref{fig:1}) to efficiently read out spiking data from the chip.
Using these spiking data we were able to learn 3D trajectories in a noise-robust way (Figure \ref{fig:6}C).
The pooling layer successfully increased the simulation speed to faster than real-time.
It was also intended to make the spiking activity more invariant to small changes in the network, which is the exact purpose of using pooling layers in deep neural networks \citetext{\citealp[Chapter~9.3]{goodfellow2016deep}; \citealp{boureau2010theoretical}}.
A pooling layer has been applied to spiking neural networks before \citep{tavanaei2017bio, tavanaei2019deep}, but -- to the best of our knowledge -- such a structure has never been applied to enhance the performance of read-outs from recurrent network architectures.
The fact that the pooling layer improved performance for the anisotropic network in our study indicates that implementing pooling layers in reservoir computing architectures could be useful in other cases, for example when the reservoir has spatially-dependent connectivity \citep{maass2002real}, and especially for reducing parameters on algorithms running on neuromorphic hardware.


Taken together, in this study we provide an algorithm for storing stable trajectories in spiking neural networks, optimized for the neuromorphic hardware Loihi.
The network architecture is capable of executing these trajectories on demand in real-time given noisy, and even never-before-seen, inputs.
While an exhaustive exploration of the parameter space remains the subject of future work, we have shown that the anisotropic network admits stable sequences with sufficient variability for output learning across hundreds of milliseconds, making it suitable for applications reaching far beyond motor control.
Further, we demonstrated that spike-based pooling can implement on-chip regularisation for the ansiotropic network, improving read out speed and accuracy.
In contrast, in the randomly connected network nearby neurons show uncorrelated activity and spatial pooling has no benefit.
Thus, spatial pooling in locally-connected SNNs proved to be a promising feature, specifically for real-time robotic control on neuromorphic hardware.
Importantly, we provide the first neuromorphic implementation which has no global learning or adaptation mechanism and produces noise-robust spiking patterns on a control-relevant timescale with sufficient variability to learn arbitrary functions.

While other approaches employing spiking neural networks exist, in general they fail to meet at least one of the mentioned criteria.
This means, in their current form, these models are either not implementable on neuromorphic hardware or do not produce sequences that are stable, variable and long enough.
We briefly describe these models and highlight how they may be adapted for neuromorphic implementation.

%

\citet{laje2013robust} presented an ``innate training'' approach. 
The network was initialized with a short input pulse and a modified FORCE algorithm \citep{sussillo2009generating} was used to train the recurrent connections.
This stabilizes the innate structure of the recurrent connections and allows a network state between chaotic and locally stable activity patterns.
A trained output trajectory was robust to perturbations, due to the tuned recurrent weights.
Unfortunately this algorithm uses a rate coded network and non-local learning rules, both of which are not applicable for most neuromorphic systems.

\citet{pehlevan2018flexibility} analyzed different approaches to solve the stability-variability trade-off in the context of songbird songs.
One additional and important criterion for their evaluation was the ability of an algorithm to provide temporal flexibility, such that outputs can be replayed faster or slower.
They concluded that a synfire chain model fits best to solve this task. 
While this approach seems to model the dynamics underlying songbird songs with flexible timing, synfire chains have a feed-forward structure which makes them less flexible than recurrent network types.

\citet{hennequin2014optimal} put more focus on getting stable output from unstable initial conditions.
They used an optimization algorithm to build an inhibitory structure that helps to stabilize the excitatory activity.
More precisely, the strength of existing inhibitory connections was changed or new inhibitory synapses were created or removed using an algorithm based on a relaxation of the spectral abscissa of the weight matrix \citep{vanbiervliet2009smoothed}.
With this they obtained relatively stable spiking dynamics.
Interestingly, this approach is similar to our study in a sense that both approaches focus on the weight matrix.
While their proposed solution to the stability-variability trade-off is promising, so far the algorithm has mainly been tested with rate coded networks.
A more elaborate analysis with a spiking neural network would be of interest.

Another recent approach involves multiplexing oscillations in a spiking neural network \citep{vincent2020learning,miall1989storage}.
Two input units inject sine-waves into a reservoir of neurons and the spiking dynamics in the reservoir follow a stable and unique pattern, which enables the learning of a long and stable output.
Compared to our algorithm, the oscillating units provide a continuous input to the network.
We see this approach as a potential alternative to the anisotropic network for robotic control.
Interestingly, stability is encoded in time rather than space, which raises the question whether this approach could be combined with a pooling layer, reflecting temporal structure instead of spatial structure.

\cite{maes2020learning} trained a recurrently connected spiking network such that small groups of neurons become active in succession and thus provide the basis for a simple index code.
Via a supervisor signal, output neurons are trained to become responsive to a particular group or index from the recurrent network and, thus, fire in a temporal order encoded in the feed-forward weights to the output layer.
Importantly, learning within the recurrent network and from the recurrent network to the output layer is done using spike-timing dependent plasticity.
However, as is, their implementation has a few small, but likely reconcilable, incompatibilities with the neuromorphic hardware considered here.
For example, learning and synaptic normalisation is only local to the neuron, and not to the synapse and they rely on adaptive exponential integrate and fire neurons, which are not implemented by Loihi.
With some modifications, their model may provide another neuromorphically implementable approach.



While our approach provides an algorithm for storing stable trajectories, our two-chip Loihi system is limited in the number of neurons available, constrained mainly by the high number of synapses in our recurrent network.
Since this limitation is mainly caused by the current \texttt{NxSDK} software and not by hardware, we expect an improvement in upcoming releases.
With more neurons available we expect even better stability, reducing the last remaining variations in our predictions and allowing even longer movement actions, beyond $2$ seconds.
At this point, further investigation of how performance depends on network size, network parameters, and pooling layer configuration will be of interest.

With more neurons available, one could add multiple inputs to the network.
We hypothesize that nearby input locations lead to similar activity patterns, while input regions far from each other produce distinct activity patterns.
This behavior could be used to train multiple trajectories from different input locations.
With this, more complex robotic control tasks could be performed, beyond the generation of single trajectories.

One general hurdle in developing neuromorphic implementations is the difficulty in transferring existing spiking neural network models from CPU-based implementations to neuromorphic hardware.
As outlined in the \nameref{section:methods} section, Loihi provides a fixed hardware-implemented neuron model.
It is possible to adjust parameters, but not the neuron model itself.
Therefore, a perfect match between traditional simulators like \NEST~\citep{gewaltig2020nest} or \texttt{Brian2} \citep{stimberg2019brian} and neuromorphic hardware, like Loihi, is in general an issue for future neuromorphic algorithms.
Efficient methods for translating neuroscientific models to Loihi is the subject of current work.

Finally, to complete the algorithm for autonomous use cases, in which Loihi is able to control a robot independently, an on-chip output learning algorithm is vital.
This requires the implementation of an output neuron on the chip, with appropriate on-chip output weights.
It is already possible to train weights offline and transfer them to Loihi, applied for example in \texttt{Nengo Loihi} \citep{Bekolay2014, hampo2020associative}.
We expect that the on-chip regularization inherent in spatial pooling will improve the robustness of future online output learning algorithms.

\section{Conclusion}

Taken together, we developed an algorithm which can serve as a basic unit in robotic applications.
The anisotropic network structure offers stability against noisy inputs and the overall architecture, especially using the pooling layer, paves the way for further steps in the development of algorithms for neuromorphic hardware.
Our study proposes an algorithm based on \emph{intrinsic} self-stabilizing features of a well initialized anisotropic connectivity structure, which can overcome the instability problem of spiking neural networks and support robust outputs on a timescale of seconds.

\section*{Conflict of Interest Statement}

The authors declare that the research was conducted in the absence of any commercial or financial relationships that could be construed as a potential conflict of interest.

\section*{Author Contributions}

A.L. contributed the network simulations, C.M. contributed the Loihi implementation, including the Pelenet framework, C.T. acquired funding and supervised the study. All authors designed the study and reviewed the manuscript.

\section*{Funding}

The research was funded by the H2020-FETPROACT project Plan4Act (\#732266) [CM, AL, CT], by the German Research Foundation (\#419866478) [AL, CT], and by the Intel Corporation via a gift without restrictions.

\section*{Acknowledgments}

The authors are thankful to Osman Kaya for providing Kuka robot trajectories, to Arvind Kumar and Lukas Ruff for helpful discussions, and to Tristan Stöber for improving the text. All of them helped to improve the quality of this work. Furthermore, we thank the Intel Corporation for providing access to their Loihi chip.

\section*{Data Availability Statement}

The \texttt{PeleNet} framework for Loihi, which was written for this study, can be found on GitHub (\url{https://github.com/sagacitysite/pelenet/tree/neurorobotics}). The data that support the findings of this study are available from the corresponding author on request.

\bibliography{bib}{}
\bibliographystyle{apa}

\null\newpage

\section*{Supplement}

\renewcommand{\thefigure}{S1}
\begin{figure}[!ht]
    \centering
    \includegraphics[width=\textwidth]{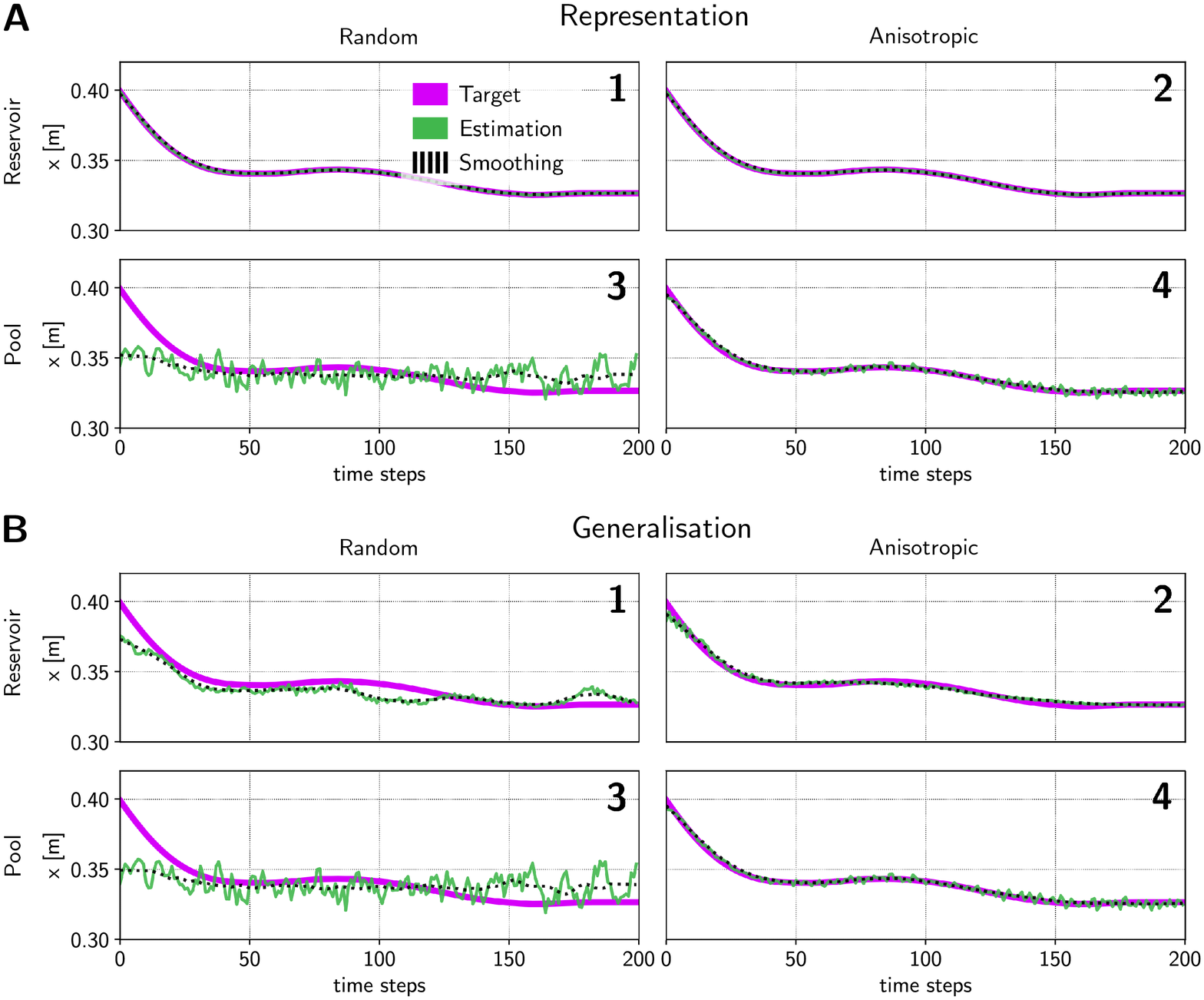}
    \caption{A single trajectory estimation for the $x$-dimension for all different tasks (representation \& generalisation), networks (randomly connected network \& anisotropic network) and estimation methods (excitatory reservoir neurons with elastic net regularization \& pooling layer neurons).}
    \label{supp:1}
\end{figure}

\null\newpage

\renewcommand{\thefigure}{S2}
\begin{figure}[!ht]
    \centering
    \includegraphics[width=\textwidth]{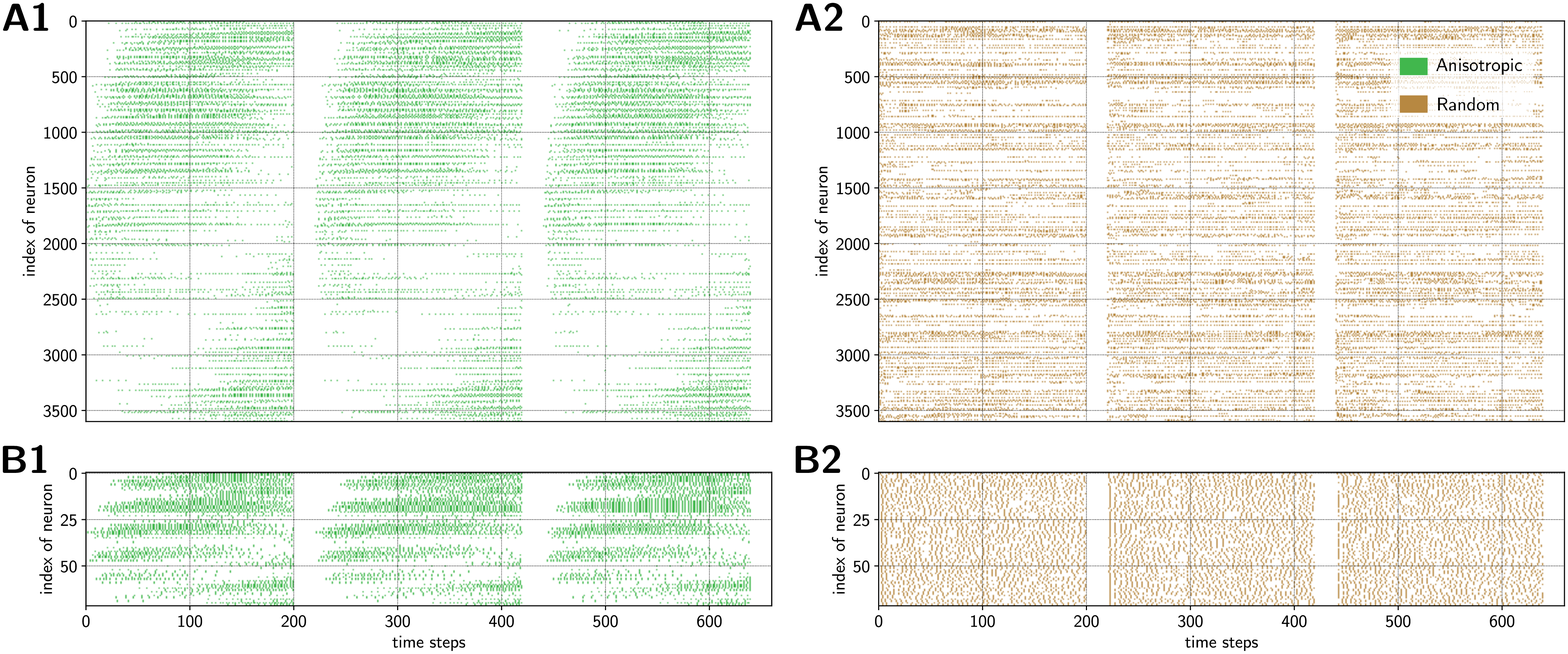}
    \caption{Spike trains of the excitatory reservoir neurons \textbf{A} compared with the spike trains of the pooling layer neurons \textbf{B}. In the anisotropic network (green) the stream-like structure of the excitatory reservoir neurons are reflected in the pooling layer.}
    \label{supp:2}
\end{figure}

\null\newpage


\renewcommand{\thetable}{S1}
\begin{table}[!ht]
\centering
\bgroup
\def\arraystretch{1.5}
\begin{tabular}{ll>{\centering\arraybackslash}p{30mm}>{\centering\arraybackslash}p{30mm}}
\hline
\multicolumn{2}{l}{Parameter} & NEST & Loihi \\ \hline
temporal resolution & $dt$ & $0.1\,ms$ & N/A \\
excitatory neurons & $npop_E$ & $3600$ & $3600$ \\
inhibitory neurons & $npop_I$ & $900$ & $900$ \\
membrane capacitance & $C_m$ & $250.0\,pF$ & N/A \\
leak conductance & $g_L$ & $25.0\,nS$ & N/A \\
threshold potential & $v_{th}$ & $-55.0\,mV$ & $64000$ \\
resting potential & $E_L$ & $-70.0\,mV$ & $0$ \\
reset potential & $v_{reset}$ & $-70.0\,mV$ & $0$ \\
refractory period & $t_{ref}$ & $2.0\,ms$ & $2$ \\
synaptic time constant (exc.) & $\tau_{exc}$ & $5.0\,ms$ & N/A \\
synaptic time constant (inh.) & $\tau_{inh}$ & $5.0\,ms$ & N/A \\
current decay  & $\tau_I$ & N/A & $380$ \\
voltage decay & $\tau_v$ & N/A & $400$ \\
synaptic delay & $d$ & $1.0\,ms$ & $1$ \\
synaptic weights (excitatory) & $J^{exc}$ & $40\,pA$ & $12$ \\
synaptic weights (inhibitory) & $J^{inh}$ & $-160\,pA$ & $48$ \\
connection probability & $p_{conn}$ & $0.05$ & $0.05$ \\
perlin scale & $\kappa_{perlin}$ & $4$ & $4$ \\
gaussian sigma (exc.) & $\sigma_E$ & $12$ & $12$ \\
gaussian sigma (inh.) & $\sigma_I$ & $9$ & $9$ \\
shift magnitude & $n_{shift}$ & $1$ & $1$ \\ \hline
\end{tabular}
\egroup
\caption{Comparison of parameters used for the \NEST~and the Loihi simulation. Both implementations use leaky integrate-and-fire neurons with current-based synapses. The \NEST~model has an additional alpha-function shaped synaptic current rise, which is not available on Loihi.}
\label{tab:1}
\end{table}

\end{document}